\documentclass{svproc}
\usepackage{comment}
\usepackage[utf8]{inputenc} % allow utf-8 input
\usepackage[T1]{fontenc}    % use 8-bit T1 fonts
\usepackage{hyperref}       % hyperlinks
\usepackage{url}            % simple URL typesetting
\usepackage{booktabs}       % professional-quality tables
\usepackage{amsfonts}       % blackboard math symbols
\usepackage{nicefrac}       % compact symbols for 1/2, etc.
\usepackage{microtype}      % microtypography
\usepackage{lipsum}
\usepackage{footmisc}

\usepackage{slashbox}
\usepackage[libertine]{newtxmath}
\usepackage{color}
\usepackage{setspace}
\usepackage{xcolor}
\usepackage{pdflscape}
\usepackage{graphicx} % import images
\usepackage{float}
\usepackage{algorithm}
\usepackage{algorithmicx}
\usepackage{algpseudocode}	
\usepackage{indentfirst}

\title{%LARE: A novel LAte-Fusion framework for Speaker REcognition\\
 DeepMSRF: A novel Deep Multimodal Speaker Recognition framework with Feature selection }

\titlerunning{ DeepMSRF: A novel Deep Multimodal Speaker Recognition framework with Feature selection}

\author{Ehsan Asali \and   Farzan Shenavarmasouleh \and   Farid Ghareh Mohammadi  \and Prasanth Sengadu Suresh \and Hamid R. Arabnia   }
\authorrunning{Ehsan Asali et al.}
  
  \institute{ Department of Computer Science,\\ University of Georgia, Athens, Georgia, 30602
   \email{ehsanasali@uga.edu,farzan.shenavarmasouleh@uga.edu farid.ghm@uga.edu,ps32611@uga.edu,hra@uga.edu}}

\begin{document}

\maketitle
\begin{abstract}
  %Speaker recognition has always been a challenging problem for the deep learning and machine learning communities.
  For recognizing speakers in video streams, significant research studies have been made to obtain a rich machine learning model by extracting high-level speaker's features such as facial expression, emotion, and gender. However, generating such a model is not feasible by using only single modality feature extractors that exploit either audio signals or image frames, extracted from video streams. In this paper, we address this problem from a different perspective and propose an unprecedented multimodality data fusion framework called \textbf{DeepMSRF}, Deep Multimodal Speaker Recognition with Feature selection. We execute DeepMSRF by feeding features of the two modalities, namely speakers' audios and face images. DeepMSRF uses a two-stream VGGNET to train on both modalities to reach a comprehensive model capable of accurately recognizing the speaker's identity. We apply DeepMSRF on a subset of VoxCeleb2 dataset with its metadata merged with VGGFace2 dataset. The goal of DeepMSRF is to identify the gender of the speaker first, and further to recognize his or her name for any given video stream. 
  The experimental results illustrate that DeepMSRF outperforms single modality speaker recognition methods with at least 3 percent accuracy. 
  
\end{abstract}

%%%%%%%%%%%%%%%%%%%%%%%%%%%%%%%%%%%%%%%%%%%%%%%%%%%%%%%%%%%%%%%%%%%%%%%%%%%%%%%%

\section{INTRODUCTION}

Artificial Intelligence (AI) has impacted almost all research fields in the last decades. There exist countless number of applications of AI algorithms in various areas such as medicine \cite{afshar2020taste}\cite{sotoodeh2019improving}, robotics \cite{buffinton2020investigating}\cite{haeri2019thermodynamics}\cite{seraj2020coordinated}, multi-agent systems \cite{dadvar2020multiagent}\cite{etemad2020using}\cite{karimi2014mining}, and security and privacy \cite{tahmasebian2020crowdsourcing}. Deep learning is, with no doubt, the leading AI methodology that revolutionized almost all computer science sub-categories such as IoT \cite{profarabnia4DeepLearning}, Computer Vision, Robotics, and Data Science \cite{mohammadi2020introduction}. The field of Computer Vision has been looking to identify human beings, animals and other objects in single photo or video streams for at least two decades. Computer vision provides variety of techniques such as image/video recognition \cite{profarabnia3}\cite{mohammadi2019parameter}, image/video analysis, image/video segmentation \cite{wang20182d}, image/video captioning, expert's state or action recognition \cite{soanssa}, and object detection within image/video \cite{ren2015faster}\cite{shenavarmasouleh2020drdr}. Object detection plays a pivotal role to help researchers find the most matching object with respect to the ground truth. The greatest challenge of object recognition task is the effective usage of noisy and imperfect datasets, especially video streams. In this paper, we aim to address this issue and propose a new framework to detect speakers.

%In addition, most algorithms do not account for the importance of feature selection in the input data streams. With the deep learning community being the latest player in the game, we have a powerful tool to work-around these limitations. Deep Learning, put in simple terms, is just a cascade of neural networks (NNs), with the predecessor feeding its processed output as the input to the next NN until a wholesome resulting model is generated. Each NN has several layers with its own hyper parameters to learn and process the data available. One such popular deep learning tool used majorly in speaker recognition and object classification is the convolutional neural networks (CNNs). The very first successful implementation of CNNs is said to be in the year 1994, however, it is said to have been in use long before that. 

\subsection{Problem Statement}
 Copious amount of research has been done to leverage single modality which is either using audio or image frames. However, very little attention has been given to multimodality based frameworks. The main problem is speaker recognition where the number of speakers are around 40. In fact, when the number of classes (speakers) proliferate to a big number and the dimension of extracted features becomes too high, traditional machine learning approaches may not be able to yield high performance due to the problem of curse of dimensionality \cite{ch1_farid}\cite{ch2_farid}. To explore the possibility of using multimodality, we feed the video streams to the proposed network and extract two modalities including audio and image frames. We aim to use feature selection techniques in two different phases in the proposed method.

Now this approach may prompt some questions: Why do we need multimodality if just single modality would give us a high enough accuracy? Is it always better to add more modalities or would an additional modality actually bring down the performance? If so, by how much? Bolstered by our experimental results, these are some questions we are going to delve into and answer in this paper. Let's start by looking at the potential impediments we could run into while using a single modality. Let's say, for instance, we just use audio-based recognition systems; in this case, we often face a bottleneck called SNR(Signal-to-Noise-Ratio) degradation, as mentioned in \cite{chetty2008robust}. In short, when SNR is low in the input dataset, we observe our model efficiency plummets. On the other hand, image-based data is not unfettered by such predicaments as well. Images face problems like pose and illumination variation, occlusion, and poor image quality \cite{mudunuri2015low}\cite{hussain2015robust}\cite{sellahewa2010image}\cite{li2018face}. Thus, we hypothesize that combining the two modalities and assigning appropriate weights to each of the input streams would bring down the error rate.\smallskip
\subsection{Feature Selection}
Feature selection is arguably one of the important steps in pre-processing before applying any machine learning algorithms. Feature selection or dimension reduction works based on two categories, including (i) filter-based and (ii) wrapper-based feature selection. Filter-based feature selection algorithms evaluate each feature independent of other features and only relies on the relation of that feature with target value or class label. This type of feature selection is cheap, as it does not apply any machine learning algorithms to examine the features. On the contrary, wrapper-based feature selection algorithms choose subsets of features and evaluate them using machine learning algorithms. That is the main reason why wrapper-based feature selection algorithms are more expensive. Mohammadi and Abadeh \cite{mohammadi2014image,mohammadi2014new} applied wrapper-based algorithms for binary feature selections using artificial bee colony. In this study, we apply wrapper based feature selection, as it yields a high performance on supervised datasets.\smallskip

\subsection{Contribution}
Most of the speaker recognition systems currently deployed are based on modelling a speaker, based on single modality information, i.e., either audio or visual features. The main contributions of this paper are as following: 
\begin{itemize}
  \item Integration of audio and image input streams extracted from a video stream, forming a multimodality deep architecture to perform speaker recognition.
  \item Effectively identifying the key features and the extent of contribution of each input stream.
  \item Creating a unique architecture that allows segregation and seamless end-to-end processing by overcoming dimensionality bottlenecks.
\end{itemize}\smallskip

\begin{figure*}[t]
    \centering
    \includegraphics[width=\textwidth]{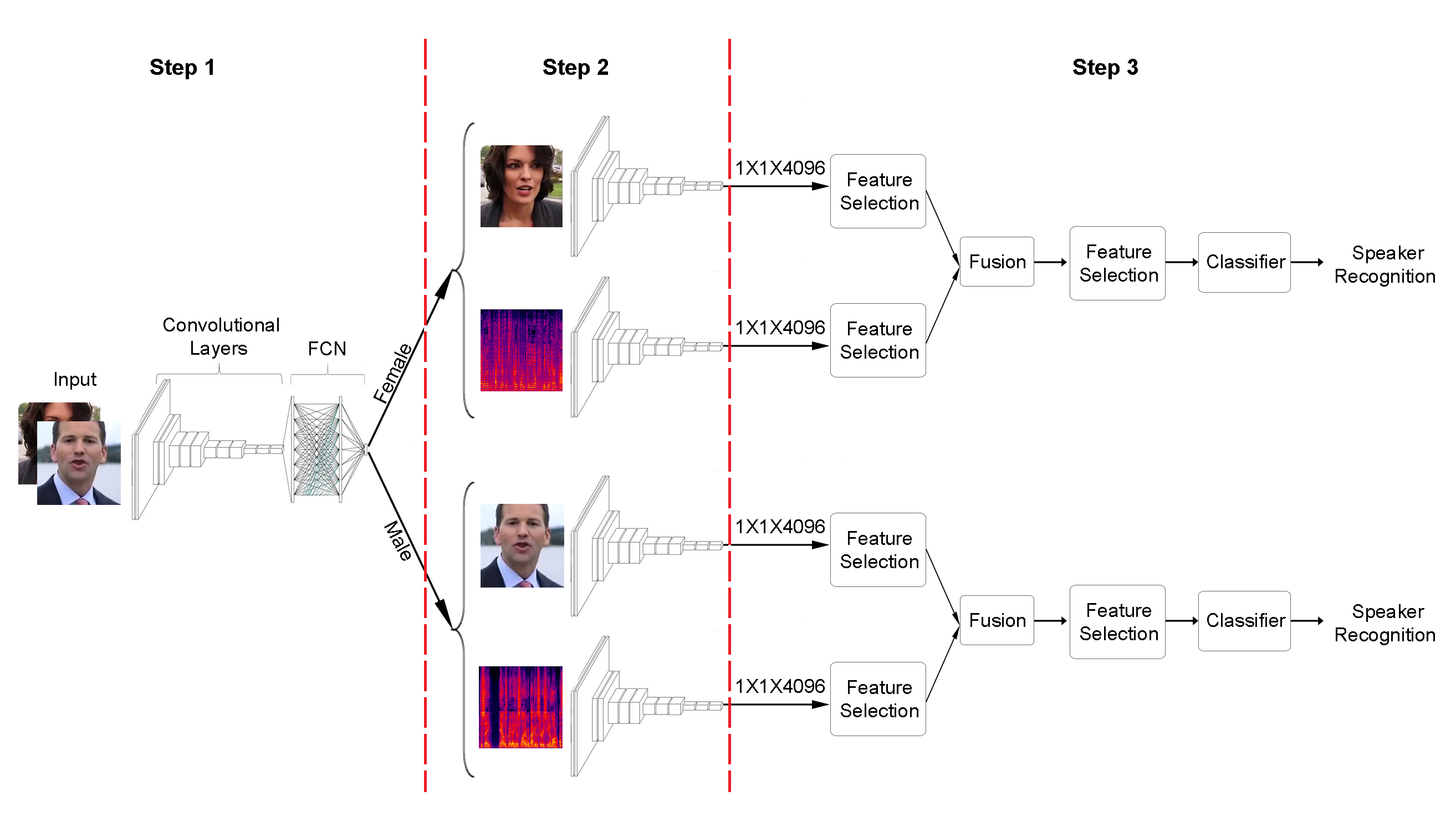}
    
    \caption{DeepMSRF architecture. Step 1: A unimodal VGGNET takes the speaker's image as an input and detects the speaker's gender. Step 2: Based on the gender, the image and voice of the speaker is passed to the corresponding parallel multimodal VGGNETs to extract each modality's dense features. Step 3: Feature selection on each modality will be applied first; then, the resultant feature vectors are concatenated, and feature selection is performed again after concatenation. Eventually, a classifier is trained to recognize the speaker's identity.}
    
\label{fig:base}
\end{figure*}

The rest of the paper is arranged as follows: First, we touch base with the related work that has been done in this field, then we explain the overview working methodology of CNNs, followed by how we handle the data effectively. We also compare and contrast other classifiers that could be used instead of the built-in neural network of VGGNET. Then, we explain the experiments performed, compare the results with some baseline performance and conclude with discussion, future work and varied applications of the model developed in this paper.

\section{RELATED WORK}
% kossher! kir dahane prasanth:  \cite{chung2018voxceleb2} uses image data to recognize the gender and identify whether the speaker is smiling or not.

As explained before, most of the work done so far on speaker recognition is based on unimodal strategies. However, with the advancement of machine learning and deep learning in the past few years, it has been proven that multimodal architectures can easily surpass unimodal designs. \cite{chetty2008robust}\cite{koda2009facial}\cite{chibelushi1994voice} were some of the very first attempts to tackle the task of person identity verification or speaker recognition while leveraging multiple streams to combine data collected from different sources such as clips recorded via regular or thermal cameras, and the varieties of corresponding features extracted from them via different external speech recognition systems, optical flow modules and much more. After the features were extracted and fused, some basic machine learning models such as Hidden Markov Model (HMM), Latent Dirichlet Allocation (LDA), and Principal Component Analysis (PCA) were trained over them to act as the final classifier.

As impressive as these look, they can never beat the accuracy that one can achieve with deep learning models. Almost all of the architectures that are at the cutting edge in the modern tasks, make use of two or more streams. Video action recognition is one example. \cite{feichtenhofer2016convolutional} and \cite{dana2017} both employ two parallel streams, one for capturing spatial data and the other for extracting temporal information from videos to classify the actions. Similarly, \cite{peng2016multi} uses two separate streams for RGB frame and sequence of different flows, whereas \cite{yang2016multilayer} brings four modalities into play and makes use of 2D and 3D CNNs and their flows simultaneously. Another excellent work \cite{soanssa} that deals with multimodal inputs, suggests a unique framework for recognizing robots' state-action pairs which uses two streams of RGB and Depth data in parallel. More creatively, \cite{feichtenhofer2019slowfast} utilizes one slow and one fast stream, proving that the former is good to understand spatial semantics, and the latter, which is a lighter pathway, can be beneficial in finding out the temporal motion features. Also, \cite{xiao2020audiovisual} builds on top of this and adds one more stream to engage audios as well. 

Tracking objects in videos, finding tampered images, and testing audio-visual correspondence (AVC) are some other tasks that parallel streams have been used for them to achieve the state of the art performance. \cite{feichtenhofer2017detect} leverages two streams to jointly learn how to detect and track objects in the videos at the same time. \cite{he2018twofold} uses two parallel Siamese networks to do real-time object tracking and \cite{zhou2017two} employs face classification and patch triplet streams to investigate the possible alterations to the face images. Also, \cite{arandjelovic2017look} and \cite{cramer2019look} both use parallel streams to enable their models to identify whether the input audio corresponds to a given video or not. \\
It can be perceived that an extensive amount of research has been done in the field of multimodal deep architectures. Nevertheless, to the best of our knowledge, \cite{dhakal2019near} is the most related work that has been done for the task of speaker recognition and it only uses multimodality in the process of feature extraction. Additionally, it only uses audio data and tests on two datasets with 22 and 26 speakers, respectively. On the contrary, in this paper, we design our architecture to make use of video frames along with the audios in separate streams. On top of that, we create our custom dataset with 40 unique speakers and extend the scale of previous works.

\section{Proposed Method}
In this section, we propose a late-fusion framework using a dual-modality speaker recognition architecture using audios and frames extracted from videos. Firstly, we discuss challenges in speaker classification and recognition, then we talk about the bottlenecks of the architecture, and finally we present our model's architectural details.

\subsection{Challenges}
As the number of images and videos proliferate, the process of image/video classification becomes more challenging; so the task of real world computer vision and data analysis becomes crucial when the number of classes exceeds 10. The more the classes, the more time and computational power is required to do the task of classification. To learn a model to classify the speakers, we are required to have a proper dataset and a structured framework to do that. In this paper, the greatest challenge was to recognize 40 unique speakers. More so than that, since no standard dataset is available for our hypothesis, we had to create ours by subsampling from a combination of two other datasets. During the last ten years, researchers have proposed different frameworks for deep learning using complex combination of neural networks such as ResNet \cite{he2016deep} and GoogleNet \cite{szegedy2015going} for image classification. These only focus on singular modality, either image or voice of the video, to do speaker recognition. In this paper, we address this problem and propose a new architecture leveraging VGGNET (VGG-16)  \cite{zhang2015accelerating} for speaker recognition task using multimodality to overcome all these limitations. The simple VGGNET, like other frameworks, suffers from having insufficient performance on speaker recognition. We provide three main steps consisting of combining two networks of VGGNET followed by feature selection and performing late-fusion on top of them. \\
Another common conundrum is on how to interpret the audio signals into a format which is suitable for VGGNET to work with. In general, in order to deal with audio streams, we have three options to choose from. One is to map the input audios to waveform images and feed the resultant diagrams to VGGNET as input. Another choice is to apply feature extraction to obtain a meaningful representation of the audio streams which is now a feature vector. The last but not the least, we can perform one more step on top of feature extraction by visualizing them as a two-dimensional image. Later in this paper we will see that the third choice has the best performance and is utilized in our final model.

\subsection{ Video Speaker Recognition }
We present a base speaker recognition architecture that leverages two VGG16 networks in parallel; One for speakers' images and another for speakers' audio frames. We discuss generating speaker audio frames in the next subsection. Figure \ref{fig:base} illustrates the base speaker recognition architecture. VGGNET produces a 1-D vector of 4,096 features for each input frame that could be used as an input to all common classification methodologies. Fusing these feature vectors yields to high dimensionality problem called curse of dimensionality (COD) \cite{ch1_farid}\cite{ch2_farid}. To reduce the problem's complexity, we apply feature selection as discussed earlier in section 1.2.

\subsubsection{Data Preparation}
We prepare a standard dataset of speaker images, along with speaker audios trimmed to four seconds, about which we discuss further in the dataset subsection under the experiment section. The dataset of the speakers' face images can be created quite easily; Nevertheless, as previously mentioned, we should convert speaker voices into proper formats as well. To tackle this issue, we have tried various available methods to generate meaningful features out of input audio signals. There are a couple of choices which we have briefly mentioned previously and will investigate more in this section. \\
The first approach is to directly convert the audio files into wave form diagrams. To create such images, the main hurdle that we face is the frequency variation of the speakers' voices. To solve this issue, we plot them with the same y-axis range to have an identical axes for all the plots. Obviously, the y-axis length must be such that all wave form charts fall within its range. The generated images can directly be fed into VGGNET; Nonetheless, later we will see that this approach is not really useful because of lacking sufficiently descriptive features.\\
Another approach is to extract meaningful and descriptive features of the audio streams instead of just drawing their waveform diagrams. Here, we have multiple options to examine; Mel-frequency cepstral coefficients (MFCCs), Differential Mel-frequency cepstral coefficients DMFCCs, and Filter Banks (F-Banks) are the algorithms reported to be effective for audio streams. MFCCs and DMFCCs can each extract a vector of 5,200 features from the input audio file, while F-Banks can extract 10,400 features. Now we have the option of either using these feature vectors directly and concatenating them with the extracted feature vectors of the face images coming from VGGNET Fully-Connected Layer 7 (so-called FC7 layer) or mapping them first to images and then, feeding them into the VGGNET. In the latter, we first need to fetch the flattened FC7 layer feature vector; afterwards, we have to perform the concatenation of the resultant vector with the previously learned face features. Spectrograms are another meaningful set of features we use in this paper. A spectrogram is a visual representation of the spectrum of frequencies of a signal (audio signal here) as it varies with time. We feed such images into VGGNET directly and extract the features from the FC7 layer. Later, we will see how beneficial each of the aforementioned approaches is to predict the speakers' identity.

\subsubsection{Feature selection}
The more features we have, the higher is the probability of occurring overfitting problems which is also known as Curse of Dimensionality. This can be resolved to some extent by making use of feature selection (FS) algorithms. Feature selection approaches carefully select the most relevant and important subset of features with respect to the classes ( speakers' identities). We choose a wrapper-based feature selection by exploiting lib-SVM kernel to evaluate the subsets of features. After applying feature selection, the dimensionality of the dataset decreases significantly. In our work, we apply FS two times in the model; once for each of the modals separately, and one again after concatenating them together (before feeding the resultant integrated feature vector to the SVM classifier.

\subsection{ Extended video Speaker Recognition }
To do the task of video speaker recognition we have divided our architecture into three main steps as presented in fig. \ref{fig:base}. The following sections explain each step in a closer scrutiny.

\subsubsection{Step One}
 Inevitably, learning to differentiate between genders is notably more straightforward for the network compared to distinguishing between the identities. The former is a binary classification problem while the latter is a multi-label classification problem with 40 labels (in our dataset). On the other hand, facial expressions and audio frequencies of the two genders differ remarkably in some aspects. For instance, a woman usually has longer haircuts, a smaller skull, jewelries, and makeup. Men, on the contrary, occasionally have a mustache and/or beard, colored ties, and tattoos. Other than facial characteristics, males mostly have deep, low pitched voices while females have high, flute-like vocals. Such differences triggered the notion of designing the first step of our framework to distinguish the speakers' genders.\\
 Basically, the objective of this step is to classify the input into Male and Female labels. This will greatly assist in training specialized and accurate models for each class. Since the gender classification is an easy task for the VGGNET, we just use the facial features in this step. In fact, we pass the speakers' images to the VGGNET and based on the resultant identified gender class extracted from the model, we decide whether to use the network for Male speakers or Female ones. In our dataset, such a binary classification yields 100 $\%$ accuracy on the test set. Later we will see the effect of this filtering step on our results.
 
 %To do so, we apply VGGNET on two sets of speaker datasets (images and voices) to extract 4096 features.  These extracted features make two datasets. Then, we apply fully connected NN to evaluate the performance of the generated datasets. These datasets only include two classes: one man, one female. It means that the VGGNET is capable of distinguishing between men and women. Thus we need to keep this first filtering and move on to second step and seek speakers' names.
 
 \subsubsection{Step Two}
 In this step, we take the separated datasets of men and women as inputs to the networks. For each category, we apply two VGGNETs, one for speaker images and another for their voices. Thus, in total, we have five VGGNETs in the first and second steps. Indeed, we had one singular modality VGGNET for filtering out the speakers' genders in the first step, and we have two VGGNETs for each gender (four VGGNETs in total) in the second step. Note that the pipeline always uses the VGGNET specified for the gender recognition in the first step; Afterwards, based on the output gender, it chooses whether to use the parallel VGGNET model for women or men, but not both simultaneously.\\

In the given dataset, we have 20 unique females and 20 unique males. Following this, we train the images and audios of males and females separately on two parallel VGGNETs and extract the result of each network's FC7 layer. Each extracted feature vector consists of 4,096 features which is passed to the next step.\\

The second step may change a little if we use the non-visualized vocal features of either MFCCs, DMFCCs, or F-Bank approach. In this case, we only need VGGNET for the speakers' face images to extract the dense features; Following this, we concatenate the resultant feature vector with the one we already have from vocal feature extractors to generate the final unified dataset for each gender.

\subsubsection{Step Three}
After receiving the feature vectors for each modality of each gender, we apply a classifier to recognize the speaker. Since the built-in neural network of VGGNET is not powerful enough for identity detection, we try a couple of classifiers on the resultant feature vector of the previous step to find the best classifier for our architecture. Nonetheless, before we feed the data into the classifiers, we need to ensure the amount of contribution each modality makes to the final result. As the contribution of each modality on the final result can vary according to the density and descriptivity of its features, we need to filter out the unnecessary features from each modality. To do so, we apply feature selection on each modality separately to allocate appropriate number of features for each of them. Afterwards, we concatenate them together as a unified 1-D vector. We apply feature selection again on the unified vector and use the final selected features as input to the classifiers. The specific number of data samples, the number of epochs for each stage and the results at each checkpoint are discussed in the Experiments section (section 4). 

%FC7 8192 feature vector input to svm classifier
\section{Experiments}

\subsection{Dataset}
We have used VoxCeleb2 dataset proposed in \cite{chung2018voxceleb2} which originally has more than 7,000 speakers, 2,000 hours of videos, and more than one million utterances. We use an unbiased sub-sample \cite{shenavarmasouleh2019causes} of that with 20,000 video samples in total, with almost 10,000 sample speakers per gender. The metadata of VoxCeleb2 dataset has gender and id labels; the id label is connected to the VGGFace2 dataset. The first step we have to go through is to bind the two metadata sets together and segregate the labels correctly. The way their dataset is arranged is that a celeb id is assigned to multiple video clips extracted from several YouTube videos which is almost unusable. Hence, we unfolded this design and assigned a unique id to each video to make them meet our needs. \smallskip

As mentioned earlier, we selected 40 random speakers from the dataset which included 20 male and 20 female speakers. Thereupon, one frame per video was extracted where the speaker's face was clearly noticeable. The voice was also extracted from a 4-second clip of the video. Finally, the image-voice pairs shuffled to create training, validation, and test sets of 14,000, 3,000 and 3,000 samples respectively for the whole dataset, i.e. both genders together. 
 
 \subsection{Classifiers}

\subsubsection{Random Forests}
Random forests \cite{ho1995random} or random decision forests are an ensemble learning method for classification, regression and other tasks that operates by constructing a multitude of decision trees at training time and outputting the class that is the mode of the classes (classification) or mean prediction (regression) of the individual trees. Random decision forests prevent the overfitting which is common in regular decision tree models. \smallskip

\subsubsection{Gaussian Naive Bayes}
Naïve Bayes \cite{john1995estimating} was first introduced in 1960s (though not under that name) and it is still a popular (baseline) method for classification problems. With appropriate pre-processing, it is competitive in the domain of text categorization with more advanced methods including support vector machines. It could also be used in automatic medical diagnosis and many other applications.

%Naïve Bayes classifiers are highly scalable, requiring a number of parameters linear in the number of variables (features/predictors) in a learning problem. Maximum-likelihood training can be done by evaluating a closed-form expression, which takes linear time, rather than by expensive iterative approximation as used for many other types of classifiers. \smallskip

\subsubsection{Logistic Regression}
Logistic regression is a powerful statistical model that basically utilizes a logistic function to model a binary dependent variable, while much more complicated versions exist. In regression analysis, logistic regression \cite{kleinbaum2002logistic} (or logit regression) is estimating the parameters of a logistic model. Mathematically, a binary logistic model has a dependent variable with two possible values, such as pass/fail which is represented by an indicator variable, where the two values are labeled "0" and "1". In the logistic model, the log-odds (the logarithm of the odds) for the value labeled "1" is a linear combination of one or more independent variables ("predictors"); the independent variables can each be a binary variable (two classes, coded by an indicator variable) or a continuous variable (any real value). The corresponding probability of the value labeled "1" can vary between 0 (certainly the value "0") and 1 (certainly the value "1"), hence the labeling; the function that converts log-odds to probability is the logistic function, hence the name. The unit of measurement for the log-odds scale is called a logit, from logistic unit, hence the alternative names. Some applications of logits are presented in \cite{amini2016estimating}\cite{karamicomparison}\cite{shahabiniaestimating}.
%Analogous models with a different sigmoid function instead of the logistic function can also be used, such as the probit model; the defining characteristic of the logistic model is that increasing one of the independent variables multiplicatively scales the odds of the given outcome at a constant rate, with each independent variable having its own parameter; for a binary dependent variable this generalizes the odds ratio.

\subsubsection{Support Vector Machine}
A Support Vector Machine \cite{hearst1998support} is an efficient tool that helps to create a clear boundary among data clusters in order to aid with the classification. The way this is done is by adding an additional dimension in cases of overlapping data points to obtain a clear distinction and then projecting them back to the original dimensions to break them into clusters. These transformations are called kernels. SVMs have a wide range of applications from finance and business to medicine\cite{maddah2020use} and robotics \cite{khayami2014cyrus}. An example of its applications can be found in \cite{asali2018namira}\cite{asali2016shiraz}\cite{asali2016ml} where they detect the opponent team's formation in a Soccer 2D Simulation environment using various approaches including SVM. We use linear kernel in our experiments and we have Linear-SVM as a part of our proposed pipeline. \smallskip

\begin{comment}

%\subsubsection{Neural Networks}
%A Neural Network (NN) is a bunch of algorithms which endeavors to discern underlying correlations in a set of data through a process that mimics the way the human brain operates. A “neuron” in a neural network is a mathematical function that gathers and classifies information according to a particular architecture. The network has an intensive resemblance to statistical approaches namely curve fitting and regression analysis.

%A NN enjoys layers of interconnected nodes. Each node is a perceptron which feeds the signal generated by a multiple linear regression into an activation function that may be nonlinear. NNs have a wide range of applications from finance and business to medicine\cite{maddah2020use} and robotics \cite{khayami2014cyrus, khayami2014m}.

\end{comment}

\subsection{VGGNET Architecture}
This section briefly explains the layers of our VGGNET architecture. Among the available VGGNET architectures, we have chosen the one containing the total of 13 convolutional and 3 Dense layers, famed as VGG-16 \cite{simonyan2014very}. The architecture includes an input layer of size $224\times224\times3$ equal to a 2-D image with 224 pixels width and the same height including RGB channels. The input layer is followed by two convolutional layers with 64 filters each and a max pooling layer with a window of size $2\times2$ and the stride of 2. Then another pair of convolutional layers of size $112\times112$ with 128 filters each and a max pooling layer are implemented. Afterwards, in the next three stages, the architecture uses three convolutional layers and one pooling layer at the end of each stage. The dimension of the convolutional layers for these steps are $56\times56\times256$, $28\times28\times512$, and $14\times14\times512$, respectively. Finally, it has three dense layers of size $1\times1\times4096$ followed by a softmax layer. Since the output of the softmax layer specifies the output label (e.g. the speaker's name), its size must be equal to the number of classes. Also, notice that all convolutional and dense layers are followed by a ReLU function to protect the network from having negative values. Moreover, the first Dense layer is usually referred as FC7 layer (Fully Connected layer 7) that contains an extracted flattened feature vector of the input.

\subsection{Implementation}

In section 4.1, we explained how we create the dataset and now we elucidate the steps taken to produce the results. In order to train the parallel VGGNET for each gender, we divide the dataset into two parts; the samples of the 20 Male speakers and the samples of the 20 Female speakers. Thereafter, each of the two partitions is fed into a dual-channel VGGNET consisting the image and audio streams. When the training process finishes, the architecture learns to extract meaningful features from the input data. Now, we can generate a new dataset for each gender by passing the face and Spectrogram's train, validation, and test images through their corresponding VGGNETs and fetch their FC7 layers' feature vectors. \\
Afterwards, using the linear SVM feature extractor library in Scikit learn - Python, we are able to extract almost 1,053 number of features for Male images, 798 features for Male voices, 1,165 features for Female images and 792 features for Female voices. Then, we concatenate the resultant feature vectors for each gender and feed it again to the same feature extractor to summarize it once more. The final size of the merged feature vectors for the Males is 1,262 and for Females is 1,383. Note that the reported number of voice features are related to the Spectrogram feature extractor which is the one we elected among the available options that were discussed earlier in section 3.3. The last step is to train the Linear SVM classifier and to get its result.\\

The very first baseline architecture that we are going to compare our results with, does not segregate genders, uses only one modality (i.e. either the face or the voice data), uses the plotted wave form of the voice data, and does not use any feature selection approach. To compare the effect of any changes to the baseline, we have accomplished an extremely dense ablation study process. The ablation study results are discussed in the next section.

\subsection{Ablation Study}
To check the effect of each contribution, we perform ablation study by training and testing the dataset in various conditions. The following sections briefly discuss the impact of each contribution on the final result.

\subsubsection{Feature Extraction and Selection}
Feature Extraction (FE) is highly crucial in the learning process. The main contribution of deep learning pipelines over the classical machine learning algorithms is their ability to extract rich meaningful features out of a high dimensional input. Here, VGGNET plays this role for the face images of the speakers and also for the visualized vocal features. On the other hand, Feature Selection (FS) can prevent the model to be misled by irrelevant features. As previously mentioned, we have used linear-SVM feature selection in this work.\\
To evaluate the advantage of using FE and FS when dealing with audial data and also to compare the performance of diverse FE algorithms, we apply each algorithm on Male, Female, and the whole dataset. Then, we apply FS on top of it; then, we examine each algorithm with four different classification methods, including Random Forests, Naive Bayes, Logistic Regression, and Support Vector Machines. Finally, we compare the results for both cases of either using or not using FS. Table \ref{table:VoiceFE} shows the result for all the situations. As the results represent, the best test accuracy is achieved when we utilize Spectrogram feature extractor combined with linear-SVM feature selection approach.

\begin{table}[htb]
\centering
\caption{The accuracy for single/multi modality with/without feature selection \smallskip}
\label{table:VoiceFE}
% \resizebox{\columnwidth}{!}{%
\begin{tabular}{|c|c|c|c|c|c|}

    \hline 
    \backslashbox{\textbf{FE Algorithm}}{\textbf{classifier}}  & RF & NB & LR & SVM    \\ 
         \hline

     Spectogram(M) ($\%$) &45.4 &19.06&54.33&50.53 \\ \hline
     Spectogram(M) + FS ($\%$)  &45.93&25.93&52.86&\textbf{56.26} \\ \hline
     
     Spectogram(F)($\%$)&44.26&21.53&52.26&48.46\\ \hline
      Spectogram(F) + FS ($\%$)&42.66&29.4&51.2&\textbf{53.3}\\ \hline

     Spectogram(all)($\%$)&37.16&14.96&48.4&43.6\\ \hline
     Spectogram(all) + FS ($\%$)&38.03&21.6&46.5&\textbf{49.3}\\\hline
      
    Waveform(M)($\%$)  &30.53&16.6&32.26&29.26\\ \hline
      Waveform(M)($\%$) +FS  &30.46&17.2&29.93&32.13\\ \hline
      
    Waveform(F)($\%$)&22.06&14.08&27.73&21.4\\ \hline
    Waveform(F)($\%$) +FS &22&13.93&23.26&23.6\\ \hline
    
     MFCC(M)($\%$)&11.93&25.33&5.46&9.46\\ \hline
    MFCC(M)($\%$) +FS &11.46&24.2&5.33&9.2\\ \hline  
    
     MFCC(F)($\%$)&10.8&21.86&5.93&9.46\\ \hline
    MFCC(F)($\%$) +FS &10.8&21.53&5.93&9.73\\ \hline  
    
     %DMFCC(M)($\%$)&11.93&25.33&5.46&6.4&9.46\\ \hline
    %DMFCC(M)($\%$) +FS &11.46&24.2&5.33&5.93&9.2\\ \hline  
    
     %DMFCC(F)($\%$)&10.8&21.86&5.93&6.73&9.46\\ \hline
    %DMFCC(F)($\%$) +FS &10.8&21.53&5.93&6.2&9.73\\ \hline 

      Filter bank(M)($\%$)&32.33&19.13&42.06&36.6\\ \hline
    Filter bank(M)($\%$) +FS &33&24&42.26&40.46\\ \hline  
    
     Filter bank(F)($\%$)&33.66&18.06&43.2&38.06\\ \hline
    Filter bank(F)($\%$) +FS &33&25.46&41.93&41.06\\ \hline 

\end{tabular}
    %  \small{\footnotesize{FS: Feature Selection, FE: Feature Extractor, M: Male samples accuracy, F: Female samples accuracy, LR: Logistic Regression, NB: Naive Bayes, RF: Random Forests, SVM: Support Vector Machines}}
% }
\end{table}

To analyze the efficacy of FS on the face frames, we train VGGNET and extract the FC7 layer feature vector. We then apply FS and eventually, train on four different classifiers. Table \ref{table:FaceFE} represents the test accuracy for each classifier with and without FS. As the results demonstrate, the highest accuracy for each dataset is achieved for the case in which we have used FS on top of VGGNET and for the SVM classifier.

\begin{table}[htb]
\centering
\caption{The accuracy of the speakers' face images for four different classifiers associated with different feature extractors with/without Feature Selection (FS) \smallskip  }
\label{table:FaceFE}
% \resizebox{\columnwidth}{!}{%
\begin{tabular}{|c|c|c|c|c|c|}

\hline 
    \backslashbox{\textbf{FE}}{\textbf{classifiers}}  & RF & NB & LR & SVM    \\ 
         \hline

     VGG(M) ($\%$)  &91 &49.66&93.6&91.66 \\ \hline
     VGG(M) + FS ($\%$)  &91.26&66.73&92.53&\textbf{94.2} \\ \hline
     
       VGG(F)($\%$)&86.26&55.33&90.93&87.33\\ \hline
      VGG(F) + FS ($\%$)&85.66&62.2&88.13&\textbf{91.26}\\ \hline
      VGG(total)  ($\%$)&88.03&50.1&91.53&88.7\\ \hline
      VGG(total) + FS ($\%$)&88.06&58.43&90.4&\textbf{91.9}\\ \hline

\end{tabular}
    %  \small{\footnotesize{FS: Feature Selection}}
% }
\end{table}

\subsubsection{Gender Detection}

As discussed in section 3.3.1 in details, the first step of our pipeline is to segregate speakers by their gender. Instead, we could train a model with 40 classes consisting of all men and women speakers. To see how the first step improves the overall performance of the model, we examined both cases and compared their results. The test accuracy of Male speakers, Female speakers, the average test accuracy of Male and Female speakers, and the test accuracy of the whole dataset (containing both genders) are reported in the table \ref{table:finalResult}. The results show that the average accuracy increases when we perform gender segregation regardless of whether we use feature selection before and/or after concatenating the face and audio modalities or not. Also, notice that according to the table \ref{table:finalResult}, we can achieve the highest accuracy when we perform feature selection, specifically before the concatenation step. According to the table, the average accuracy has been improved by almost 4 percent using our proposed method (the last row) compared to the baseline approach (the first row).

\begin{table}[htb]
\centering
\caption{The accuracy of the whole dataset for SVM classifier associated with Spectrogram feature extractor combined with feature selection \smallskip}
% \resizebox{\columnwidth}{!}{%
\begin{tabular}{|c|c|c|c|c|}

    \hline 
    \backslashbox{\textbf{approach}}{\textbf{samples}}  & Male & Female &Avg.& Total(Genderless)    \\ 
         \hline

      Simple concatenation &91.2&87.87&89.54&89.27\\  \hline
  FS + concatenation &\textbf{95.13}&91.87&93.5&92.97\\   \hline
  concatenation + FS &94.67&91.87&93.27&92.53\\  \hline
  FS + concatenation + FS &95.07&\textbf{91.93}&\textbf{93.5}&\textbf{93.03}\\  \hline

\end{tabular}
    % }
% \small{\footnotesize{FS: Feature Selection}}
       \label{table:finalResult}
\end{table}

\subsubsection{Single modality Vs. Multimodality}
One of the greatest contributions of DeepMSRF is taking advantage of more than one modality to recognize the speaker efficiently. Each modality comprises of unique features that lead the model to distinguish different individuals. To show how multimodality can overcome the limitations of single modality, we carry out a comparison between the two, reported in table \ref{table:modality}. According to the results, using both visual and auditory inputs together can improve the accuracy of the task of speaker recognition.

\begin{table}[htb]
\centering
\caption{The Accuracy for single/multi modality with/out feature selection \smallskip }
\label{table:modality}
% \resizebox{\columnwidth}{!}{%\smallskip 
\begin{tabular}{|c|c|c|c|}

    \hline 
    \backslashbox{\textbf{Result}}{\textbf{Modality}}  & Face Frames & Audio (Spectrogram) & Multimodality\\ 
         \hline
   %   Male ($\%$)  &88.7 &50.53&43.6 \\ \hline
    % Male + FS ($\%$)  &91.9&56.26& \\ \hline
     %female ($\%$)  &88.7 &43.6&89 \\ \hline
     %female + FS ($\%$)  &91.9&49.3&93.03 \\ \hline
     Total ($\%$)  &88.7 &43.6&89 \\ \hline
     Total + FS ($\%$)  &91.9&49.3&93.03 \\ \hline

\end{tabular}
% }
\end{table}

\subsection{Time Complexity}
\begin{figure}[ht]
    \includegraphics[width=\textwidth]{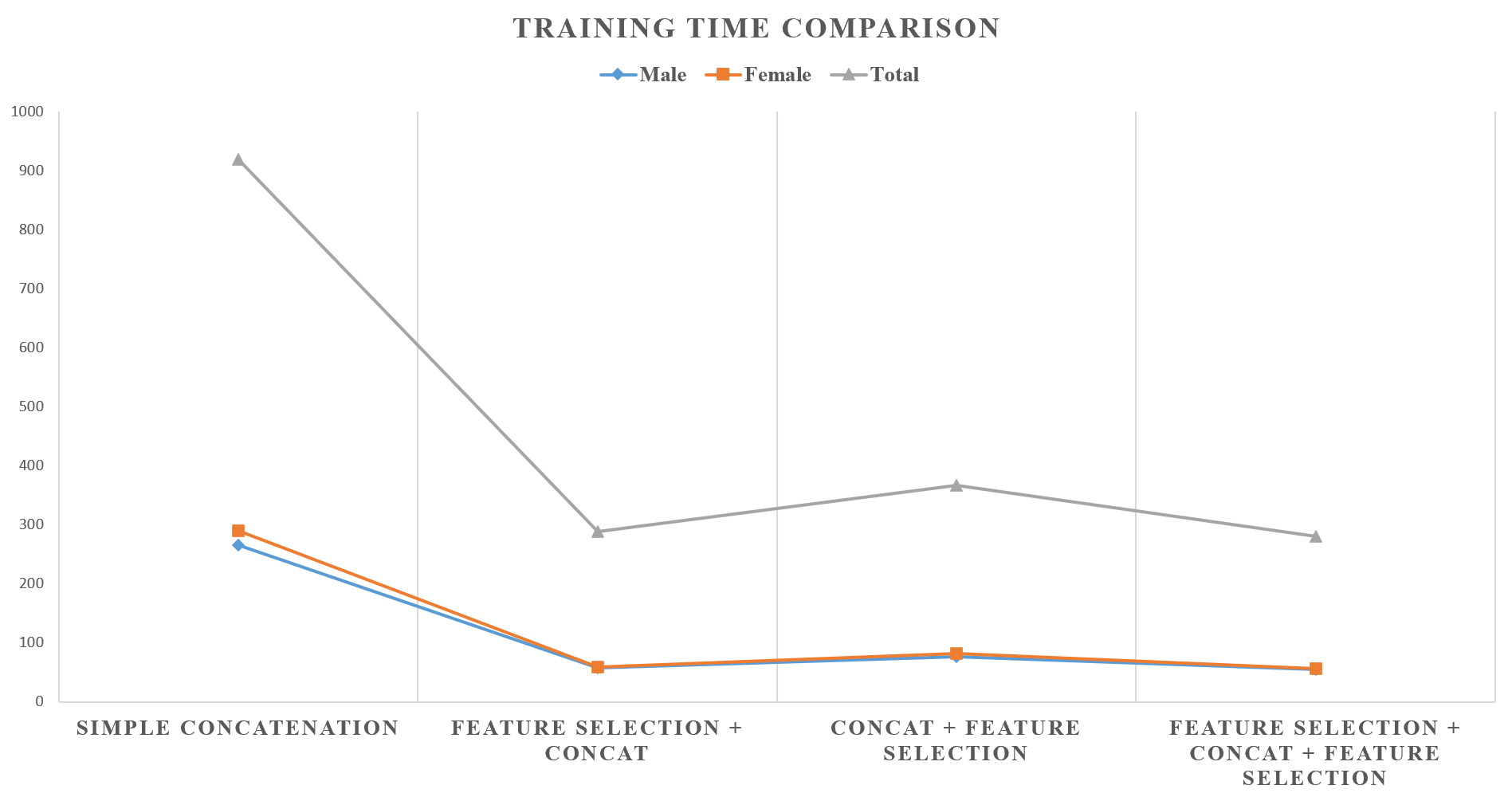}
    \caption{Training time Comparison (Male Vs. Female)
 for DeepMSRF with SVM classifier}
\label{fig:Timecomplesity}
\end{figure}

In section 4.5, we saw the benefit of utilizing feature selection on the model's accuracy. Additionally, there exist one more criteria to consider which is the training time. Although the training process is being performed offline and in the worst case, training the SVM classifier over our dataset finishes in almost 20 minutes (for the whole dataset), it is noteworthy to see how feature selection can influence the training time. Figure \ref{fig:Timecomplesity} depicts the training time required for the SVM in the last step (step3) of our pipeline for the experiments shown in table \ref{table:finalResult}. According to fig. \ref{fig:Timecomplesity}, the required training time for each gender is approximately one third of the corresponding time required to train the whole dataset. 

Nextly, the time required to accomplish the step 3 of DeepMSRF, including the time needed for feature selection and the training time, is reported in table \ref{table:finalResultTime}. The results can be discussed from two different points of view: (i) Among the examined methodologies, the least training duration is for the case in which we apply Feature Selection (FS) after the concatenation of the two modalities' feature vectors. On the contrary, the worst time performance is for the situation of applying FS before and after concatenation. (ii) The time is significantly shorter when we segregate the genders. Each gender's dataset needs less than one third of the required time for the whole dataset. In fact, training two separate models (one per gender), together, requires lesser time than training a general model which contains both genders.

\begin{table}[htb]
\centering
\caption{The accuracy of the whole dataset for SVM classifier associated with Spectrogram feature extractor combined with feature selection \smallskip}
% \resizebox{\columnwidth}{!}{%
\begin{tabular}{|c|c|c|c|}

    \hline 
    \backslashbox{\textbf{approach}}{\textbf{samples}}  & Male & Female & Total(Genderless) \\ 
         \hline

      Simple concatenation &265.37&290.94&919.46\\  \hline
  FS + concatenation &209.91&195.49&883.23\\   \hline
  concatenation + FS &179.61&197.27&729.6\\  \hline
  FS + concatenation + FS &296.93&291.56& 1,208.52\\  \hline

\end{tabular}
    % }
% \small{\footnotesize{FS: Feature Selection}}
       \label{table:finalResultTime}
\end{table}

\section{Future work}
The future work entails using multiple datasets to test the robustness of the system and adding more functionalities to the model such as recognizing the facial features of each individual person and predicting the extent of overlap between different people. Moreover, some aspects of the model can be investigated more. For instance, here we have used VGGNET; there remain a myriad of object detection pipelines to be tested. Another example of possible future probes is to try disparate approaches of feature selection. Altogether, this architecture has numerous applications beyond speaker recognition and are to be explored in the upcoming works.\\

\section{Conclusion}
This paper takes a trip down the novelty lane by adding multimodality to improve robustness of the recognition and overcomes the limitations of single modality performance.
From the results of the experiments above, we can infer that the hypothesis made about the multimodality improving over the single modality results for person recognition using deep neural networks was nearly conclusive. Among other challenges, this paper also solves the dimensionality challenge arising from using multimodality input streams. 
Exploiting feature extraction has provided a deep insight into how significant features to train the network are to be extracted to obtain a well-trained model. We can see that although the images provide a high accuracy over speaker recognition, audio stream input reinforces the performance and provides an additional layer of robustness to the model. 
In conclusion, we state that the unique framework used in this paper, DeepMSRF, provides an efficient solution to the problem of speaker recognition from video streams. At last, DeepMSRF is a highly recommended framework for those researchers who deal with video analysis.\\

\bibliographystyle{unsrt}  
\bibliography{references}  %%% Remove comment to use the external .bib file (using bibtex).

\begin{thebibliography}{10}

\bibitem{afshar2020taste}
Ardavan Afshar, Ioakeim Perros, Haesun Park, Christopher deFilippi, Xiaowei
  Yan, Walter Stewart, Joyce Ho, and Jimeng Sun.
\newblock Taste: temporal and static tensor factorization for phenotyping
  electronic health records.
\newblock In {\em Proceedings of the ACM Conference on Health, Inference, and
  Learning}, pages 193--203, 2020.

\bibitem{sotoodeh2019improving}
Mani Sotoodeh and Joyce~C Ho.
\newblock Improving length of stay prediction using a hidden markov model.
\newblock {\em AMIA Summits on Translational Science Proceedings}, 2019:425,
  2019.

\bibitem{buffinton2020investigating}
Keith~W Buffinton, Benjamin~B Wheatley, Soheil Habibian, Joon Shin, Brielle~H
  Cenci, and Amanda~E Christy.
\newblock Investigating the mechanics of human-centered soft robotic actuators
  with finite element analysis.
\newblock In {\em 2020 3rd IEEE International Conference on Soft Robotics
  (RoboSoft)}, pages 489--496. IEEE, 2020.

\bibitem{haeri2019thermodynamics}
Hossein Haeri, Kshitij Jerath, and Jacob Leachman.
\newblock Thermodynamics-inspired modeling of macroscopic swarm states.
\newblock In {\em Dynamic Systems and Control Conference}, volume 59155, page
  V002T15A001. American Society of Mechanical Engineers, 2019.

\bibitem{seraj2020coordinated}
Esmaeil Seraj and Matthew Gombolay.
\newblock Coordinated control of uavs for human-centered active sensing of
  wildfires.
\newblock {\em arXiv preprint arXiv:2006.07969}, 2020.

\bibitem{dadvar2020multiagent}
Mehdi Dadvar, Saeed Moazami, Harley~R Myler, and Hassan Zargarzadeh.
\newblock Multiagent task allocation in complementary teams: a
  hunter-and-gatherer approach.
\newblock {\em Complexity}, 2020, 2020.

\bibitem{etemad2020using}
Mohammad Etemad, Nader Zare, Mahtab Sarvmaili, Am{\'\i}lcar Soares,
  Bruno~Brandoli Machado, and Stan Matwin.
\newblock Using deep reinforcement learning methods for autonomous vessels in
  2d environments.
\newblock In {\em Canadian Conference on Artificial Intelligence}, pages
  220--231. Springer, 2020.

\bibitem{karimi2014mining}
Maryam Karimi and Marzieh Ahmazadeh.
\newblock Mining robocup log files to predict own and opponent action.
\newblock {\em International Journal of Advanced Research in Computer Science},
  5(6):1--6, 2014.

\bibitem{tahmasebian2020crowdsourcing}
Farnaz Tahmasebian, Li~Xiong, Mani Sotoodeh, and Vaidy Sunderam.
\newblock Crowdsourcing under data poisoning attacks: A comparative study.
\newblock In {\em IFIP Annual Conference on Data and Applications Security and
  Privacy}, pages 310--332. Springer, 2020.

\bibitem{profarabnia4DeepLearning}
Sahar Voghoei, Navid Hashemi~Tonekaboni, Jason Wallace, and Hamid~R Arabnia.
\newblock Deep learning at the edge.
\newblock In {\em Proceedings of International Conference on Computational
  Science and Computational Intelligence CSCI, Internet of Things" Research
  Track}, pages 895--901, 2018.

\bibitem{mohammadi2020introduction}
Farid~Ghareh Mohammadi, M~Hadi Amini, and Hamid~R Arabnia.
\newblock An introduction to advanced machine learning: Meta-learning
  algorithms, applications, and promises.
\newblock In {\em Optimization, Learning, and Control for Interdependent
  Complex Networks}, pages 129--144. Springer, 2020.

\bibitem{profarabnia3}
Soheila Amirian, Zengyan Wang, Thiab~R Taha, and Hamid~R Arabnia.
\newblock Dissection of deep learning with applications in image recognition.
\newblock In {\em Proceedings of International Conference on Computational
  Science and Computational Intelligence (CSCI 2018: December 2018, USA);
  "Artificial Intelligence" Research Track (CSCI-ISAI)}, pages 1132--1138,
  2018.

\bibitem{mohammadi2019parameter}
Farid Ghareh~Mohammadi, Hamid~R Arabnia, and M~Hadi Amini.
\newblock On parameter tuning in meta-learning for computer vision.
\newblock In {\em 2019 International Conference on Computational Science and
  Computational Intelligence (CSCI)}, pages 300--305. IEEE, 2019.

\bibitem{wang20182d}
Zengyan Wang, Fangyu Li, Thiab Taha, and Hamid Arabnia.
\newblock 2d multi-spectral convolutional encoder-decoder model for geobody
  segmentation.
\newblock In {\em 2018 International Conference on Computational Science and
  Computational Intelligence (CSCI)}, pages 1193--1198. IEEE, 2018.

\bibitem{soanssa}
Nihal Soans, Ehsan Asali, Yi~Hong, and Prashant Doshi.
\newblock Sa-net: Robust state-action recognition for learning from
  observations.
\newblock In {\em IEEE International Conference on Robotics and Automation
  (ICRA)}, pages 2153--2159, 2020.

\bibitem{ren2015faster}
Shaoqing Ren, Kaiming He, Ross Girshick, and Jian Sun.
\newblock Faster r-cnn: Towards real-time object detection with region proposal
  networks.
\newblock In {\em Advances in neural information processing systems}, pages
  91--99, 2015.

\bibitem{shenavarmasouleh2020drdr}
Farzan Shenavarmasouleh and Hamid~R Arabnia.
\newblock Drdr: Automatic masking of exudates and microaneurysms caused by
  diabetic retinopathy using mask r-cnn and transfer learning.
\newblock {\em arXiv preprint arXiv:2007.02026}, 2020.

\bibitem{ch1_farid}
Farid Ghareh~Mohammadi and M~Hadi Amini.
\newblock {E}volutionary computation, optimization and learning algorithms for
  data science.
\newblock In {\em Optimization, Learning and Control for Interdependent Complex
  Networks}. Springer, 2019.

\bibitem{ch2_farid}
Farid Ghareh~Mohammadi and M~Hadi Amini.
\newblock Applications of nature-inspired algorithms for dimension {R}eduction:
  Enabling efficient data analytics.
\newblock In {\em Optimization, Learning and Control for Interdependent Complex
  Networks}. Springer, 2019.

\bibitem{chetty2008robust}
Girija Chetty and Michael Wagner.
\newblock Robust face-voice based speaker identity verification using
  multilevel fusion.
\newblock {\em Image and Vision Computing}, 26(9):1249--1260, 2008.

\bibitem{mudunuri2015low}
Sivaram~Prasad Mudunuri and Soma Biswas.
\newblock Low resolution face recognition across variations in pose and
  illumination.
\newblock {\em IEEE transactions on pattern analysis and machine intelligence},
  38(5):1034--1040, 2015.

\bibitem{hussain2015robust}
Jamal Hussain~Shah, Muhammad Sharif, Mudassar Raza, Marryam Murtaza, and Saeed
  Ur-Rehman.
\newblock Robust face recognition technique under varying illumination.
\newblock {\em Journal of applied research and technology}, 13(1):97--105,
  2015.

\bibitem{sellahewa2010image}
Harin Sellahewa and Sabah~A Jassim.
\newblock Image-quality-based adaptive face recognition.
\newblock {\em IEEE Transactions on Instrumentation and measurement},
  59(4):805--813, 2010.

\bibitem{li2018face}
Pei Li, Loreto Prieto, Domingo Mery, and Patrick Flynn.
\newblock Face recognition in low quality images: A survey.
\newblock {\em arXiv preprint arXiv:1805.11519}, 2018.

\bibitem{mohammadi2014image}
Farid Ghareh~Mohammadi and M~Saniee Abadeh.
\newblock Image steganalysis using a bee colony based feature selection
  algorithm.
\newblock {\em Engineering Applications of Artificial Intelligence}, 31:35--43,
  2014.

\bibitem{mohammadi2014new}
Farid Ghareh~Mohammadi and Mohammad~Saniee Abadeh.
\newblock A new metaheuristic feature subset selection approach for image
  steganalysis.
\newblock {\em Journal of Intelligent \& Fuzzy Systems}, 27(3):1445--1455,
  2014.

\bibitem{koda2009facial}
Yasunari Koda, Yasunari Yoshitomi, Mari Nakano, and Masayoshi Tabuse.
\newblock A facial expression recognition for a speaker of a phoneme of vowel
  using thermal image processing and a speech recognition system.
\newblock In {\em RO-MAN 2009-The 18th IEEE International Symposium on Robot
  and Human Interactive Communication}, pages 955--960. IEEE, 2009.

\bibitem{chibelushi1994voice}
Claude~C Chibelushi, F~Deravi, and JS~Mason.
\newblock Voice and facial image integration for person recognition.
\newblock 1994.

\bibitem{feichtenhofer2016convolutional}
Christoph Feichtenhofer, Axel Pinz, and Andrew Zisserman.
\newblock Convolutional two-stream network fusion for video action recognition.
\newblock In {\em Proceedings of the IEEE conference on computer vision and
  pattern recognition}, pages 1933--1941, 2016.

\bibitem{dana2017}
Dana Rezazadegan, Sareh Shirazi, Ben Upcroft, and Michael Milford.
\newblock Action recognition: From static datasets to moving robots.
\newblock 01 2017.

\bibitem{peng2016multi}
Xiaojiang Peng and Cordelia Schmid.
\newblock Multi-region two-stream r-cnn for action detection.
\newblock In {\em European conference on computer vision}, pages 744--759.
  Springer, 2016.

\bibitem{yang2016multilayer}
Xiaodong Yang, Pavlo Molchanov, and Jan Kautz.
\newblock Multilayer and multimodal fusion of deep neural networks for video
  classification.
\newblock In {\em Proceedings of the 24th ACM international conference on
  Multimedia}, pages 978--987, 2016.

\bibitem{feichtenhofer2019slowfast}
Christoph Feichtenhofer, Haoqi Fan, Jitendra Malik, and Kaiming He.
\newblock Slowfast networks for video recognition.
\newblock In {\em Proceedings of the IEEE International Conference on Computer
  Vision}, pages 6202--6211, 2019.

\bibitem{xiao2020audiovisual}
Fanyi Xiao, Yong~Jae Lee, Kristen Grauman, Jitendra Malik, and Christoph
  Feichtenhofer.
\newblock Audiovisual slowfast networks for video recognition.
\newblock {\em arXiv preprint arXiv:2001.08740}, 2020.

\bibitem{feichtenhofer2017detect}
Christoph Feichtenhofer, Axel Pinz, and Andrew Zisserman.
\newblock Detect to track and track to detect.
\newblock In {\em Proceedings of the IEEE International Conference on Computer
  Vision}, pages 3038--3046, 2017.

\bibitem{he2018twofold}
Anfeng He, Chong Luo, Xinmei Tian, and Wenjun Zeng.
\newblock A twofold siamese network for real-time object tracking.
\newblock In {\em Proceedings of the IEEE Conference on Computer Vision and
  Pattern Recognition}, pages 4834--4843, 2018.

\bibitem{zhou2017two}
Peng Zhou, Xintong Han, Vlad~I Morariu, and Larry~S Davis.
\newblock Two-stream neural networks for tampered face detection.
\newblock In {\em 2017 IEEE Conference on Computer Vision and Pattern
  Recognition Workshops (CVPRW)}, pages 1831--1839. IEEE, 2017.

\bibitem{arandjelovic2017look}
Relja Arandjelovic and Andrew Zisserman.
\newblock Look, listen and learn.
\newblock In {\em Proceedings of the IEEE International Conference on Computer
  Vision}, pages 609--617, 2017.

\bibitem{cramer2019look}
Jason Cramer, Ho-Hsiang Wu, Justin Salamon, and Juan~Pablo Bello.
\newblock Look, listen, and learn more: Design choices for deep audio
  embeddings.
\newblock In {\em ICASSP 2019-2019 IEEE International Conference on Acoustics,
  Speech and Signal Processing (ICASSP)}, pages 3852--3856. IEEE, 2019.

\bibitem{dhakal2019near}
Parashar Dhakal, Praveen Damacharla, Ahmad~Y Javaid, and Vijay Devabhaktuni.
\newblock A near real-time automatic speaker recognition architecture for
  voice-based user interface.
\newblock {\em Machine Learning and Knowledge Extraction}, 1(1):504--520, 2019.

\bibitem{he2016deep}
Kaiming He, Xiangyu Zhang, Shaoqing Ren, and Jian Sun.
\newblock Deep residual learning for image recognition.
\newblock In {\em Proceedings of the IEEE conference on computer vision and
  pattern recognition}, pages 770--778, 2016.

\bibitem{szegedy2015going}
Christian Szegedy, Wei Liu, Yangqing Jia, Pierre Sermanet, Scott Reed, Dragomir
  Anguelov, Dumitru Erhan, Vincent Vanhoucke, and Andrew Rabinovich.
\newblock Going deeper with convolutions.
\newblock In {\em Proceedings of the IEEE conference on computer vision and
  pattern recognition}, pages 1--9, 2015.

\bibitem{zhang2015accelerating}
Xiangyu Zhang, Jianhua Zou, Kaiming He, and Jian Sun.
\newblock Accelerating very deep convolutional networks for classification and
  detection.
\newblock {\em IEEE transactions on pattern analysis and machine intelligence},
  38(10):1943--1955, 2015.

\bibitem{chung2018voxceleb2}
Joon~Son Chung, Arsha Nagrani, and Andrew Zisserman.
\newblock Voxceleb2: Deep speaker recognition.
\newblock {\em arXiv preprint arXiv:1806.05622}, 2018.

\bibitem{shenavarmasouleh2019causes}
Farzan Shenavarmasouleh and Hamid~R. Arabnia.
\newblock Causes of misleading statistics and research results
  irreproducibility: A concise review.
\newblock In {\em 2019 International Conference on Computational Science and
  Computational Intelligence (CSCI)}, pages 465--470. IEEE, 2019.

\bibitem{ho1995random}
Tin~Kam Ho.
\newblock Random decision forests.
\newblock In {\em Proceedings of 3rd international conference on document
  analysis and recognition}, volume~1, pages 278--282. IEEE, 1995.

\bibitem{john1995estimating}
George~H John and Pat Langley.
\newblock Estimating continuous distributions in bayesian classifiers.
\newblock In {\em Proceedings of the Eleventh conference on Uncertainty in
  artificial intelligence}, pages 338--345. Morgan Kaufmann Publishers Inc.,
  1995.

\bibitem{kleinbaum2002logistic}
David~G Kleinbaum, K~Dietz, M~Gail, Mitchel Klein, and Mitchell Klein.
\newblock {\em Logistic regression}.
\newblock Springer, 2002.

\bibitem{amini2016estimating}
PARSA~V AMINI, AR~SHAHABINIA, HR~Jafari, O~Karami, and A~Azizi.
\newblock Estimating conservation value of lighvan chay river using contingent
  valuation method.
\newblock 2016.

\bibitem{karamicomparison}
Omid Karami, Saeed Yazdani, Iraj Saleh, Hamed Rafiee, and Andisheh Riahi.
\newblock A comparison of zayandehrood river water values for agriculture and
  the environment.
\newblock {\em River Research and Applications}.

\bibitem{shahabiniaestimating}
Amir~Reza Shahabinia, Vahid~Amini Parsa, Hamidreza Jafari, Saeed Karimi, and
  Omid Karami.
\newblock Estimating the recreational value of lighvan chay river uses
  contingent valuation method.

\bibitem{hearst1998support}
Marti~A. Hearst, Susan~T Dumais, Edgar Osuna, John Platt, and Bernhard
  Scholkopf.
\newblock Support vector machines.
\newblock {\em IEEE Intelligent Systems and their applications}, 13(4):18--28,
  1998.

\bibitem{maddah2020use}
Erfan Maddah and Borhan Beigzadeh.
\newblock Use of a smartphone thermometer to monitor thermal conductivity
  changes in diabetic foot ulcers: a pilot study.
\newblock {\em Journal of Wound Care}, 29(1):61--66, 2020.

\bibitem{khayami2014cyrus}
Rauf Khayami, Nader Zare, Maryam Karimi, Payman Mahor, Ardavan Afshar,
  Mohammad~Sadegh Najafi, Mahsa Asadi, Fatemeh Tekrar, Ehsan Asali, and Ashkan
  Keshavarzi.
\newblock Cyrus 2d simulation team description paper 2014.
\newblock In {\em RoboCup 2014 Symposium and Competitions: Team description
  papers}, 2014.

\bibitem{asali2018namira}
Ehsan Asali, Farzin Negahbani, Saeed Tafazzol, Mohammad~Sadegh Maghareh,
  Shahryar Bahmeie, Sina Barazandeh, Shokoofeh Mirian, and Mahta Moshkelgosha.
\newblock Namira soccer 2d simulation team description paper 2018.
\newblock {\em RoboCup 2018}, 2018.

\bibitem{asali2016shiraz}
Ehsan Asali, Mojtaba Valipour, Ardavan Afshar, Omid Asali, MohammadReza
  Katebzadeh, Saeed Tafazol, Alireza Moravej, Suhair Salehi, Hosain Karami, and
  Mahsa Mohammadi.
\newblock Shiraz soccer 2d simulation team description paper 2016.
\newblock In {\em RoboCup 2016 Symposium and Competitions: Team Description
  Papers, Leipzig, Germany}, 2016.

\bibitem{asali2016ml}
Ehsan Asali, Mojtaba Valipour, Nader Zare, Ardavan Afshar, MohammadReza
  Katebzadeh, and GH~Dastghaibyfard.
\newblock Using machine learning approaches to detect opponent formation.
\newblock In {\em 2016 Artificial Intelligence and Robotics (IRANOPEN)}, pages
  140--144. IEEE, 2016.

\bibitem{simonyan2014very}
Karen Simonyan and Andrew Zisserman.
\newblock Very deep convolutional networks for large-scale image recognition.
\newblock {\em arXiv preprint arXiv:1409.1556}, 2014.

\end{thebibliography}

\end{document}